\documentclass[10pt,journal]{IEEEtran} 
\usepackage{ifpdf}

\usepackage{multirow}

\usepackage{cite}

\ifCLASSINFOpdf
  \usepackage[pdftex]{graphicx}
   \DeclareGraphicsExtensions{.pdf,.jpeg,.png, .eps, .pdf}
\else
   \usepackage[dvips]{graphicx}
   \DeclareGraphicsExtensions{.eps, .pdf}
\fi
\usepackage[cmex10]{amsmath}
\usepackage{amsfonts}
\usepackage{array}
\usepackage{fixltx2e}

\usepackage{stfloats}
\usepackage{url}

\begin{document}
%
\title{Sparse Representation-based Image Quality Assessment} 
%
%
%

\author{Tanaya~Guha,~\IEEEmembership{Student Member,~IEEE,}
        Ehsan~Nezhadarya,~\IEEEmembership{Student Member,~IEEE,}
        and~Rabab~K~Ward,~\IEEEmembership{Fellow,~IEEE}
\thanks{The authors are with the Image and Signal Processing Lab, department
of Electrical and Computer Engineering, the University of British Columbia, Vancouver,
BC, Canada.}
\thanks{email: \{tanaya, ehsann, rababw\}@ece.ubc.ca}}

%
%

\markboth{Journal of \LaTeX\ Class Files,~Vol.~6, No.~1, January~2007}%
{Shell \MakeLowercase{\textit{et al.}}: Bare Demo of IEEEtran.cls for Journals}
%



\maketitle

\begin{abstract}
A successful approach to image quality assessment involves comparing the structural information between a distorted and its reference image. However, extracting structural information that is perceptually important to our visual system is a challenging task. This paper addresses this issue by employing a sparse representation-based approach and proposes a new metric called the \emph{sparse representation-based quality} (SPARQ) \emph{index}. The proposed method learns the inherent structures of the reference image as a set of basis vectors, such that any structure in the image can be represented by a linear combination of only a few of those basis vectors. This sparse strategy is employed because it is known to generate basis vectors that are qualitatively similar to the receptive field of the simple cells present in the mammalian primary visual cortex \cite{nature}. The visual quality of the distorted image is estimated by comparing the structures of the reference and the distorted images in terms of the learnt basis vectors resembling cortical cells. Our approach is evaluated on six publicly available subject-rated image quality assessment datasets. The proposed SPARQ index consistently exhibits high correlation with the subjective ratings on all datasets and performs better or at par with the state-of-the-art.

\end{abstract}

\begin{IEEEkeywords}
Dictionary learning, Image quality, sparse representation, structural similarity.
\end{IEEEkeywords}

%

\section{Introduction}
%
%
%
%
\IEEEPARstart{D}{igital} images incur a variety of distortions during the process of image acquisition, compression, transmission, storage or reconstruction. These often degrade the visual quality of images. In order to monitor, control and improve the quality of images produced at the various stages, it is important to \emph{automatically} quantify the image quality. Since the end-users of the majority of image-based applications are humans, this requires the understanding of human perception of image quality, and mimicking it as closely as possible.

The \emph{mean squared error} (MSE) and the \emph{peak signal-to-noise ratio} (PSNR) have been traditionally used to measure the image quality degradations. These metrics, although mathematically convenient, fail to correlate well with human perception \cite{girod}. A considerable amount of research effort has been put towards quantifying the quality of images as perceived by humans, and a number of \emph{objective} image quality assessment algorithms that agree with the subjective judgment of human beings have been developed. The objective quality assessment methods, depending on whether or not they use some or all the information about the original undistorted image, are broadly classified into three categories: \emph{no-reference}, \emph{reduced-reference} and \emph{full-reference} \cite{Winkler05}. This paper concentrates on the full-reference quality estimation approach.

The earlier focus of full-reference image quality assessment research has been on building a comprehensive and accurate model of the \emph{human visual system} (HVS) and its psychophysical properties, such as the contrast sensitivity function. In this approach, the errors between the distorted and the reference images are quantized and pooled according to the HVS properties \cite{wangBook}. These methods require precise knowledge of the viewing conditions and are computationally demanding. Despite this complexity, the \emph{HVS modeling-based} methods can only make linear or quasilinear approximations of the highly non-linear HVS. Our current understanding of the HVS is also limited in many aspects. Consequently, these methods are not highly superior to MSE or PSNR \cite{msvd}. 
\begin{figure*}[t]
	\centering
		\includegraphics[width=1.0\linewidth, trim=0cm 0cm 1cm 0cm]{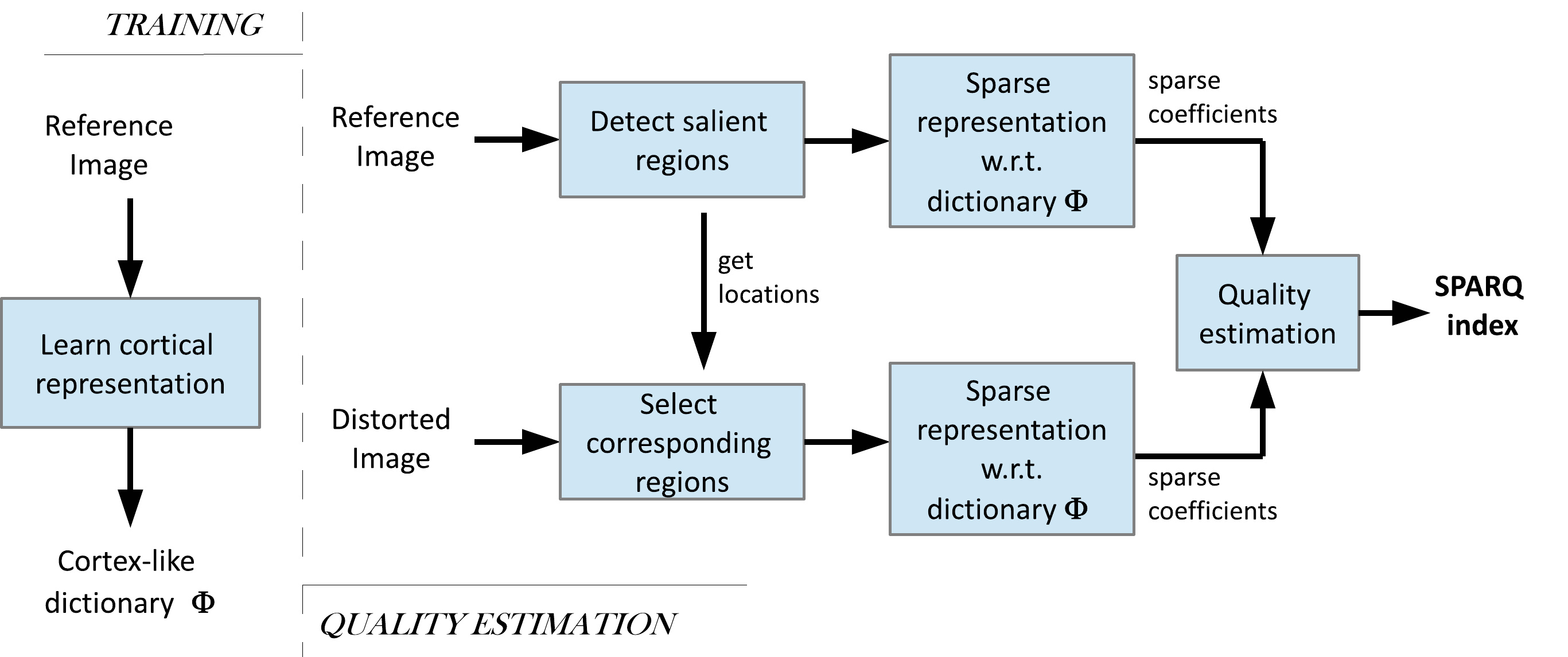}
	\caption{Overview of the proposed image quality assessment approach}
	\label{fig:overview}
\end{figure*}

The interest of modern image quality estimation research lies in modeling the content of the images based on certain significant properties of the HVS. This \emph{visual fidelity-based} approach is more attractive because of its practicality and mathematical foundation \cite{vsnr, Qsurvey}. The majority of these fidelity-based methods attempt to quantify the perceptual quality either in terms of \emph{statistical information} \cite{ifc, vif} or in terms of \emph{structural information} of the images \cite{ssim,msvd,ms-ssim,gssim,iwssim,cwssim}. The statistical approaches hypothesize that the HVS has evolved over the years to extract information from natural scenes and therefore, use natural scene statistics to estimate the perceptual quality of images. The structural approaches on the other hand operate on the basis of a rather important aspect of the HVS - its sensitivity towards the image structures for developing cognitive understanding. In this approach, image quality is estimated in terms of the \emph{fidelity of structures} between the reference and the distorted images.

The representative image quality metric of the class of structural information-based metrics is the \emph{structural similarity index} (SSIM) \cite{ssim}. SSIM treats the non-structural distortions (such as, luminance and contrast change) separately from the structural distortions. The quality of a patch in the distorted image is measured by comparing it with the corresponding patch in the original image in terms of three components: luminance, contrast and structure. A global quality score is computed by combining the effects of the three components over all image patches. SSIM achieved much success because of its simplicity, and its ability to tackle a wide variety of distortions. Due to its pixel-domain implementation, SSIM is highly sensitive to geometric distortions like scaling, translation, rotation and other misalignments \cite{wangBook}. To improve the performance of SSIM, multiscale extension \cite{ms-ssim}, wavelet transform-based modification \cite{cwssim}, gradient-domain implementation \cite{gssim} and various pooling strategies \cite{iwssim, pooling} have been proposed. 

The underlying assumption behind utilizing the structural information is that the HVS uses the structures extracted from the viewing field for its cognitive understanding. Therefore, a high-quality image is expected to preserve all the structural information of the reference image. From this viewpoint, efficiently capturing the structural information of images is the key to developing successful image quality assessment algorithms. But extracting the structural information in a perceptually meaningful way is a non-trivial task. A widely used mathematical tool for analyzing image structures is the wavelet transform. Its basis elements, being spatially localized, oriented and of bandpass in nature, resemble the receptive field of simple cells in the mammalian primary visual cortex (also known as the striate cortex and V1) \cite{nature, wangBook}. However, the wavelet transform uses a set of predefined, data-independent basis functions - the success of which is often limited by the degree as to how suitable they are in capturing the structure of the signals under consideration. 

We consider a more generalized approach to analyzing signal structures. This involves \emph{learning} a set of basis elements that are adapted to represent the inherent structures of the signal in question. These learnt basis elements are collectively known as a \emph{dictionary}. Such learning can be accomplished by fitting a set of basis vectors to a collection of training samples. As each basis vector is tailored to represent a significant part of the structures present in the given data, a learnt dictionary is more efficient in capturing the structural information compared to a predefined set of bases.

More importantly, this approach empowers us to build a \emph{cortex-like representation} of an image. In 1996, Olshausen and Field have shown that \emph{basis elements that resemble the properties of the receptive field of simple cells in the primary visual cortex can be learnt from the input images} \cite{nature}. They showed that the keys to building such a cortex-like dictionary are: (i) a \emph{sparsity prior} - an assumption that it is possible to describe the input image using a small number of basis elements, and (ii) \emph{overcompleteness} - the number of basis elements in the dictionary is greater than the vector space spanned by the input. Until recently, this important result was not exploited to its full strength in the field of signal or image processing. In the last few years, several practical dictionary learning algorithms have been developed \cite{ksvd1,mod}. It has been shown that the data-dependent, learnt dictionaries, due to their superior ability to model the inherent structures in the data, can outperform predefined dictionaries like wavelets in several image processing tasks \cite{ksvd1,ksvdenoise,restore}.

In this paper, we develop a full-reference image quality assessment metric which we name the \emph{sparse representation-based quality (SPARQ) index}. The metric relies on capturing the inherent structures of the reference image in a perceptually meaningful way. To achieve this, an overcomplete dictionary and its corresponding sparse representation are learnt from local patches of the image. The local structures in the distorted image are decomposed using the basis vectors of the learnt dictionary and the resulting sparse coefficients are used to quantify the perceptual quality of the distorted image with respect to the reference image. As our method analyzes the image structures by building a cortex-like model of the stimuli, the extracted information is expected to be perceptually meaningful. This is much different from existing structural information-based methods which, although successful, provide no evidence on the perceptual importance of the structural information they extract from images. To evaluate the efficacy of the proposed metric, we perform various experiments on six publicly available, subject-rated image quality assessment datasets: LIVE \cite{live}, A57 \cite{a57}, CSIQ \cite{csiq}, MICT \cite{toyoma} and WIQ \cite{wiq}. The proposed SPARQ index consistently exhibits high correlation with the subjective scores and often outperforms its competitors.

The rest of the paper is organized as follows. Section \ref{sec:proposed} describes the proposed quality estimation approach, followed by the experimental results and discussions in Section \ref{sec:experiments}. Section \ref{sec:conclusion} concludes the article and suggests possible directions to future work.

\section{The Proposed Approach}
\label{sec:proposed}
Our image quality assessment method is divided into two phases: a \emph{training} phase and a \emph{quality estimation} phase. The goal of the training phase is to model the inherent structures of the reference image in a perceptually meaningful way. This is achieved by learning an overcomplete dictionary from the reference image. In the quality estimation phase, a quality score, namely the SPARQ index, is computed by comparing the information in selected regions of the reference image with those in the distorted image. Figure \ref{fig:overview} presents an overview of the proposed method, and the steps are described below in detail.
\subsection{Training Phase}
\label{subsec:train}
This step involves learning (i) a dictionary i.e.\ set of basis vectors whose properties resemble those of the receptive field of simple cells in primary visual cortex, and (ii) the weights by which these basis elements are mixed together. 
\subsubsection{Motivation behind learning a cortex-like dictionary}
\label{subsubsec:cortical}
The motivation of this approach comes from the very process of image formation and how is it perceived by the HVS. The natural viewing field is highly structured and spatially correlated. The light rays that reflect off various structures in the viewing field, get focused onto an array of photoreceptors present in the retina. The information is then encoded in the form of complex statistical dependencies among the photoreceptor activities \cite{v1}. The goal of primary visual cortex, as indicated in several seminal studies \cite{nature, v1}, is to reduce these statistical dependencies in order to discover the intrinsic structures that gave rise to the image.

A reasonable strategy towards mimicking this phenomena is to describe an image in terms of a linear superposition of a few basis vectors. These basis vectors form a subset of an overcomplete set of basis vectors (dictionary) that are adapted to the given image so as to best represent the structures in the images \cite{nature, v1}. It has been shown that on employment of this strategy, the basis elements that emerge are qualitatively similar to the receptive field of the cortical simple cells \cite{nature}. The conjecture that sparsity is an important prior is based on the observation that natural images contain sparse structures and can be described by a small number of structural primitives like lines and edges \cite{v1, field94}. Due to overcompleteness, the basis vectors are also non-orthogonal and the input-output relationship deviates from being purely linear. The justification of deviating from a strictly linear approach is to account for a weak form of nonlinearity exhibited by the simple cells themselves \cite{v1}. 
%
\subsubsection{Learning a dictionary}
\label{subsubsec:v1model}
Given a reference image, $I_{ref}\in\mathbb{R}^N$, we intend to learn an overcomplete dictionary. This can be achieved by fitting the basis vectors in the dictionary to represent the local structures of the image. 

To account for the local structures in an image, a large number of distinct, possibly overlapping patches of dimension $\sqrt{n}\times \sqrt{n}$ are extracted \emph{randomly} from $I_{ref}$. Ideally, one patch centerd at every pixel should be extracted; but in practice, extracting any large number of patches is sufficient for learning a good dictionary. After extracting a large number of random patches, the patches with low or no structural information i.e.\ the homogeneous patches are discarded. This is done by removing the patches whose variance is zero or close to zero after mean removal. A number of $k$ patches are then selected from the set of the informative patches. Each image patch is converted to a vector of length $n$. These patches are concatenated to form a matrix $\mathbf{P}\in \mathbb{R}^{n\times k}$ where $k$ is the number of patches extracted from $I_{ref}$ and the columns of $\mathbf{P}$ are the patch vectors. From these patches, a dictionary $\mathbf{\Phi}=\left\{\phi_i\right\}_{i=1}^m$, $\phi_i\in\mathbb{R}^n$ is learnt. We are interested in the \emph{overcomplete} case where $n<m$ i.e.\ when $\mathbf{\Phi}$ has more basis vectors than the dimensionality of the input. An overcomplete dictionary offers greater flexibility in representing the essential structures in a signal. It is also robust to additive noise, occlusion and small translation \cite{lewicki}.

However, greater difficulties arise with overcompleteness, because a full-rank, overcomplete $\mathbf{\Phi}$ creates an underdetermined system of linear equations having an infinite number of solutions. To narrow down the choice to one well-defined solution, an additional constraint of sparsity is enforced. Let, the sparse representation of $\mathbf{P}$ over the dictionary $\mathbf{\Phi}$ be denoted by $\mathbf{X}=\left\{\mathbf{x}_i\right\}_{i=1}^k$, $\mathbf{x}_i\in\mathbb{R}^m$ where any patch vector in $\mathbf{P}$ can be represented by a linear superposition of no more than $\tau$ dictionary columns where $\tau<<m$. This is formally written as the following optimization problem:
\begin{equation}
	\begin{aligned}
	& \underset{\left\{\mathbf{\Phi}, \mathbf{X}\right\}}{\mathrm{min}}
	& & \left\{\left\Vert\mathbf{P}-\mathbf{\Phi}\mathbf{X}\right\Vert_F^2\right\}
	& \mathrm{subject~to}
	& & \left\|\mathbf{x}\right\| _{0}\leq \tau
	\end{aligned}
	\label{eq:dict1}
\end{equation}
where $\left\|.\right\|_F$ is the Frobenius norm (square root of the sum of the squared values of all elements in a matrix) and $\left\|.\right\|_0$ is the $\ell_0$ semi-norm that counts the number of non-zero elements in a vector. Although the $\ell_0$ norm provides a straightforward notion of sparsity, it renders the problem non-convex. Thus obtaining an accurate solution of \eqref{eq:dict1} is NP hard. Nevertheless, in the last few years researchers have found practical and stable ways to solve such underdetermined systems via convex optimization \cite{bp} and greedy pursuit algorithms \cite{omp}.

To solve \eqref{eq:dict1}, a recently developed learning algorithm, known as the K-SVD \cite{ksvd1} is employed. K-SVD iteratively solves \eqref{eq:dict1} by performing two steps at each iteration: (i) sparse coding and (ii) dictionary update. In the sparse coding step, $\mathbf{\Phi}$ is kept fixed and the coefficients in $\mathbf{X}$ are computed by a greedy algorithm called the orthogonal matching pursuit (OMP) \cite{omp}. 
\begin{equation}
	\begin{aligned}
	& \underset{\mathbf{X}}{\mathrm{min}}
	& & \left\{\left\Vert\mathbf{P}-\mathbf{\Phi}\mathbf{X}\right\Vert_F^2\right\}
	& \mathrm{subject~to}
	& & \left\|\mathbf{x}\right\| _{0}\leq \tau
	\end{aligned}
	\label{eq:ksvd1}
\end{equation}
In the dictionary update step, each basis element $\phi_i \in \mathbf{\Phi}$ is updated sequentially, allowing the corresponding coefficients in $\mathbf{X}$ to change as well. Updating an element $\phi_i$ involves computing a rank-one approximation of a residual matrix $\mathbf{E}_i$.
\begin{equation}
	\mathbf{E}_i = \widetilde{\mathbf{Y}_i} - \widetilde{\mathbf{\Phi}_i}\widetilde{\mathbf{X}_i}
	\label{eq:ksvd2}
\end{equation}
where $\widetilde{\mathbf{\Phi}_i}$ and $\widetilde{\mathbf{X}_i}$ are formed by removing the $i$-th column from $\mathbf{\Phi}$ and the $i$-th row from $\mathbf{X}$, and $\widetilde{\mathbf{Y}_i}$ contains only those columns of $\mathbf{Y}$ that use $\phi_i$ for their approximation. The rank-one approximation is computed by subjecting $\mathbf{E}_i$ to a Singular Value Decomposition (SVD). For the details of this learning algorithm, please refer to the original K-SVD paper \cite{ksvd1}.

\subsection{The Quality Estimation Phase}
\label{subsec:quality}
This part of our method first compares the reference and the distorted images locally, and then yields a global value as the measure of perceptual quality of the distorted image. This is accomplished through the following steps:
\begin{figure*}[tb]
	\begin{minipage}[b]{1.0\linewidth}
	\centering
		\begin{tabular}{cccc}
			\includegraphics[width=0.22\linewidth, trim=24cm 8cm 24cm 14cm, clip=true]{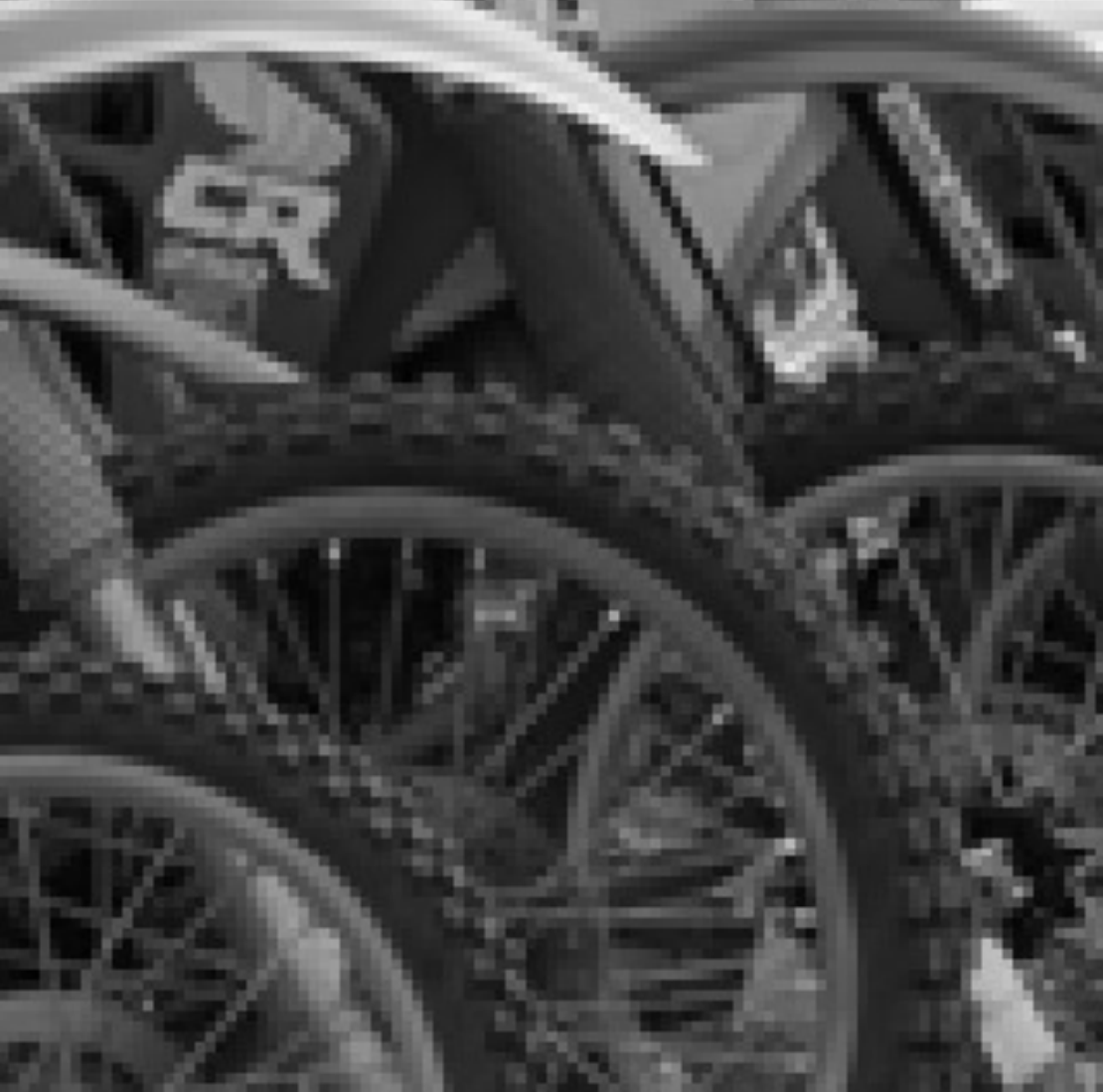} &
			\includegraphics[width=0.22\linewidth, trim=24cm 8cm 24cm 14cm, clip=true]{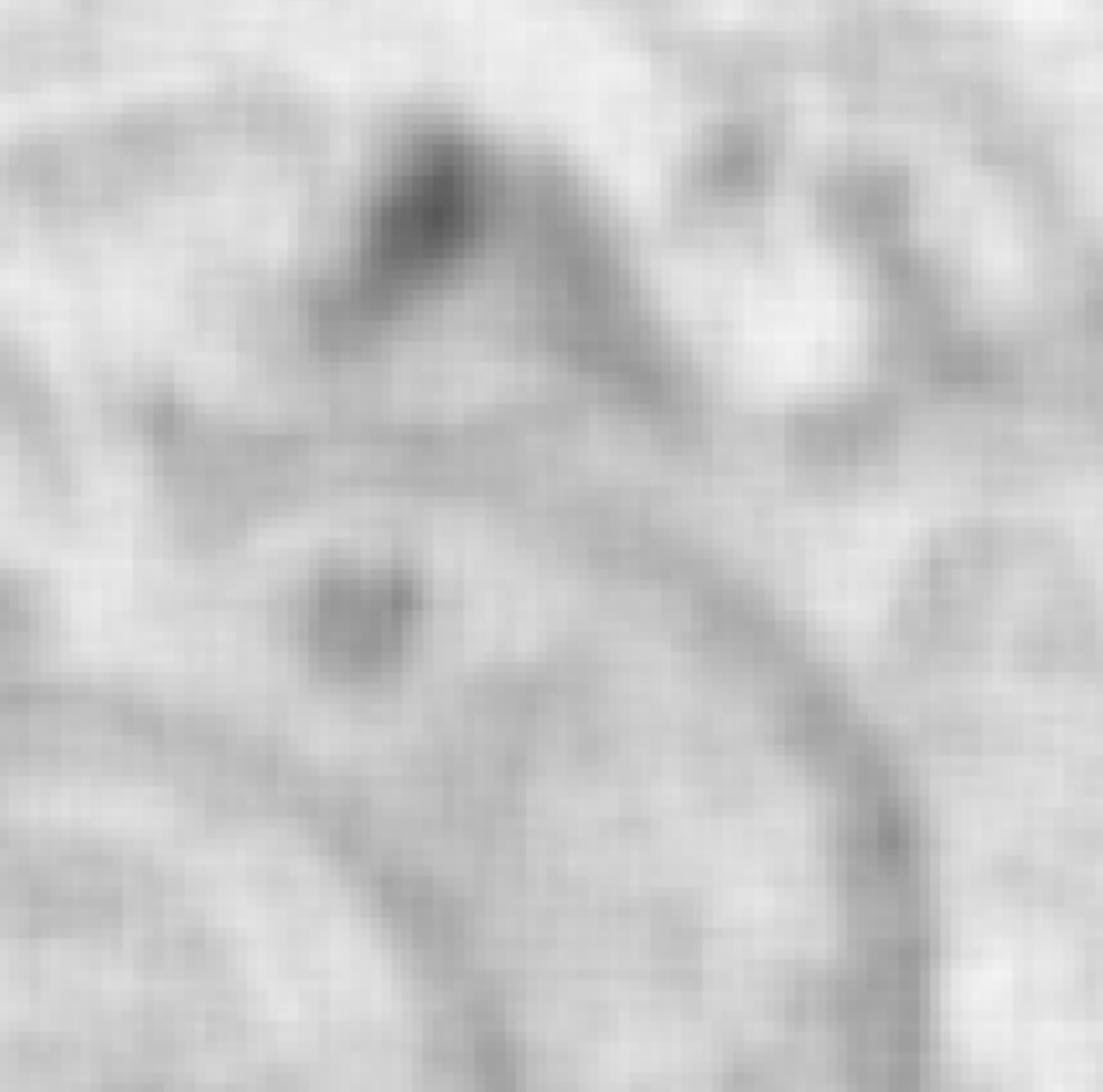} &
			\includegraphics[width=0.22\linewidth, trim=24cm 8cm 24cm 14cm, clip=true]{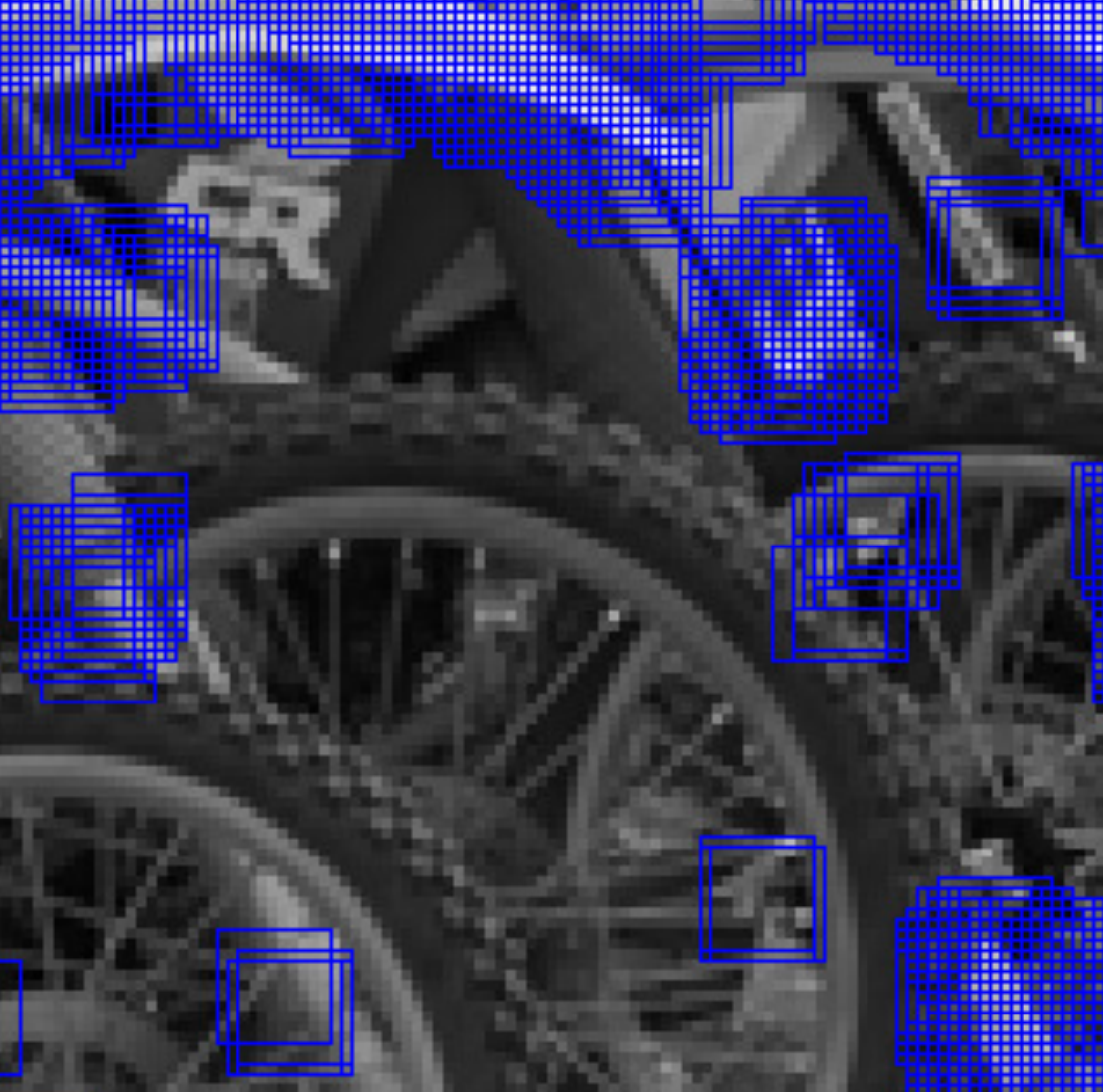} &
			\includegraphics[width=0.22\linewidth, trim=24cm 8cm 24cm 14cm,, clip=true]{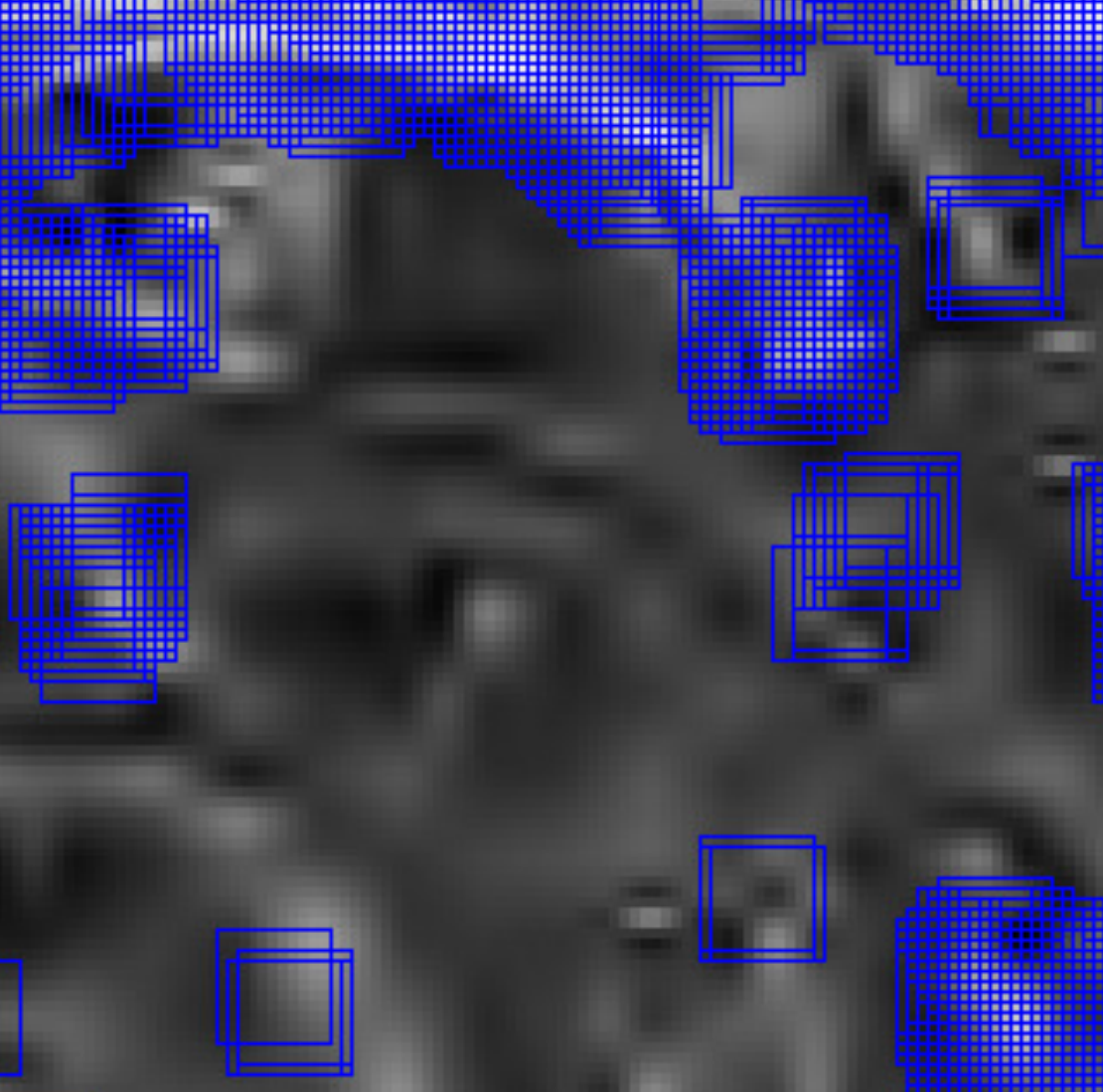} \tabularnewline
	\footnotesize (a)  &  \footnotesize (b)  & \footnotesize (c)  &  \footnotesize (d) \tabularnewline
\end{tabular}
\end{minipage}
\caption{Detection of salient regions: (a) Reference image, (b) Local entropy map of the reference image (brighter pixel value indicates higher entropy), (c) Salient patches detected in the reference image based on the entropy map, and (d) Corresponding patches in the distorted image. The images are cropped at the middle for display (best viewed in color).}
\label{fig:saliency}
\end{figure*}
\subsubsection{Detection of salient patches}
\label{subsubsec:entropy}
It is well-known that not every pixel (or region) in an image receives the same level of visual attention. Several studies have shown that significant improvement in performance of the quality metrics can be achieved by incorporating information about visual attention i.e.\ by detecting perceptually important regions \cite{larson2008, larson2008can, engelke2008regional}. 

A common hypothesis is that the HVS is an efficient extractor of information, and therefore the image regions that contain high information attract more visual attention \cite{pooling, iwssim}. Based on this hypothesis, we take an information theoretic approach towards detecting the visually important regions or patches. One way to quantify the local information content of an image is by computing the Shannon's entropy of each patch. The information content or entropy of a discrete random variable $\mathbf{z}$ with probability distribution $\mathbb{P}_z =\left\{p_1, p_2, ..., p_J\right\}$ is defined as
\begin{equation}
	H\left(\mathbf{z}\right) = H\left(\mathbb{P}_z\right)= - \sum_{j=1}^J p_j\log_2p_j
	\label{eq:entropy}
\end{equation}
Similarly, an image patch can also be analyzed as a random variable. Let us consider an image patch $\mathbf{z}$ of dimension $\sqrt{n}\times \sqrt{n}$ where each pixel in $\mathbf{z}$ is independent and identically distributed. If $\mathbf{z}$ contains $J$ distinct intensity values, its probability distribution, $\mathbb{P}_z$, is given by $\mathbb{P}_z = \left\{p_1, p_2, ..., p_J\right\}$, where $J\leq2^8$ for an 8-bit grayscale image; $p_j$ is the probability of the pixel intensity value $j$. The probability $p_j$ is defined as $p_j = f_j/n$, where $f_j$ is the number of pixels (frequency) with intensity value $j$ occurs in the image patch $\mathbf{z}$ and $n$ is the total number of pixels in $\mathbf{z}$. The entropy of every  $\sqrt{n}\times \sqrt{n}$ patch (a patch around every pixel) in the reference image $I_{ref}\in\mathbb{R}^N$ is computed as
\begin{equation}
	H\left(\mathbf{z}\right) = - \sum_{j=1}^J p_j\log_2p_j = - \frac{1}{n}\sum_{j=1}^J f_j\log_2\left(f_j/n\right)
	\label{eq:localentropy}
\end{equation}
The larger the value of $H$, the higher is the information content of a patch. 

A number of $q$ patches having the highest entropy values are selected as the \emph{salient patches} in $I_{ref}$. These patches are vectorized and arranged as columns of a matrix $\mathbf{P}_{r}\in\mathbb{R}^{n\times q}$. The locations of these $q$ patches are used to extract the corresponding patches from the distorted image $I_{dis}\in\mathbb{R}^N$. The matrix containing the patches from the distorted image is denoted as $\mathbf{P}_{d}\in\mathbb{R}^{n\times q}$. An example of this process is provided in Fig.~\ref{fig:saliency} which shows a reference image, its local entropy map, the salient patches selected in the reference image and the corresponding patches selected in the distorted image.
\subsubsection{Computation of the SPARQ index}
\label{subsubsec:sparq}
At this point, we have two sets of corresponding salient patches $\mathbf{P}_r$ and $\mathbf{P}_d$ extracted from the same locations of the reference and the distorted images. The next task is to analyze and compare these structures (patches) w.r.t. the previously learnt dictionary $\mathbf{\Phi}$. 

Let us consider a patch vector $\mathbf{p}_r\in\mathbf{P}_r$ from $I_{ref}$ and its corresponding patch vector $\mathbf{p}_d\in\mathbf{P}_d$ from $I_{dis}$. The patches $\mathbf{p}_r$ and $\mathbf{p}_d$ are decomposed using $\mathbf{\Phi}$ to obtain their respective sparse coefficients $\mathbf{x}_r$ and $\mathbf{x}_d$.
\begin{equation}
	\begin{aligned}
	& \underset{\mathbf{x}_{r}}{\mathrm{min}}
	& & \left\{\left\Vert\mathbf{p}_{r}-\mathbf{\Phi}\mathbf{x}_{r}\right\Vert_2^2\right\}
	& \mathrm{subject~to}
	& & \left\|\mathbf{x}_{r}\right\| _{0}\leq \tau
	\end{aligned}
	\label{eq:sparse1}
\end{equation}
\begin{equation}
	\begin{aligned}
	& \underset{\mathbf{x}_d}{\mathrm{min}}
	& & \left\{\left\Vert\mathbf{p}_d-\mathbf{\Phi}\mathbf{x}_d\right\Vert_2^2\right\}
	& \mathrm{subject~to}
	& & \left\|\mathbf{x}_d\right\| _{0}\leq \tau
	\end{aligned}
	\label{eq:sparse21}
\end{equation}
Note that, each of $\mathbf{x}_{r}$ and $\mathbf{x}_{d}$ contains only $\tau$ non-zero elements. The locations (indices) of these non-zero coefficients indicate those specific basis vectors in $\mathbf{\Phi}$ which actually contribute to the approximation of the input patch. These active basis vectors are called the \emph{support} of the input. The amplitudes of these non-zero coefficients are the weights by which these support vectors are combined. The support vectors and their weights together are indicative of the structural and non-structural distortions between the two input patches. Ideally, these two patches would have different sets of support vectors whenever there exist any structural distortions between them. Otherwise, if the two patches undergo purely non-structural distortions, the supports would remain the same but their weights may change.

In order to quantify the perceptual quality of $\mathbf{p}_d$ w.r.t.~$\mathbf{p}_r$, we compare their sparse representations $\mathbf{x}_d$ and $\mathbf{x}_r$. A simple but effective way to compare two vectors is to compute their \emph{normalized correlation coefficient}. A parameter $\alpha$ is computed based on the correlation coefficient between $\mathbf{x}_{r}$ and $\mathbf{x}_{d}$ as follows:
\begin{equation}
	\begin{aligned}
		\alpha(\mathbf{p}_r,\mathbf{p}_d) = \frac{\left|\mathbf{x}_r^T \mathbf{x}_d\right| + c}{\left\|\mathbf{x}_r\right\|_2 \left\|\mathbf{x}_d\right\|_2 + c}
	\end{aligned}
	\label{eq:ang}
\end{equation}
where $c$ is a small positive constant added to avoid instability when the denominator is close to zero. Clearly, $0<\alpha\leq 1$. When $\mathbf{x}_{r}$ and $\mathbf{x}_{d}$ are orthogonal, $\left|\mathbf{x}_r^T \mathbf{x}_d\right|=0$; but due to the presence of $c$, the parameter $\alpha$ is slightly greater than zero. Due to normalization, $\alpha$ is unaffected by the lengths of $\mathbf{x}_{r}$ and $\mathbf{x}_{d}$. Thus $\alpha$ is not be able to measure non-structural distortions caused by multiplying the patch elements by a constant. 

To account for these types of distortions as well, we introduce another parameter. An important measure of similarity (or difference) between two vectors is their pointwise difference. Hence, we compute another quantity $\beta$ which uses the length of the vector $\left(\mathbf{x}_r - \mathbf{x}_d\right)$.
\begin{equation}
	\begin{aligned}
		\beta(\mathbf{p}_r,\mathbf{p}_d) = 1 - \frac{\left\|\mathbf{x}_r - \mathbf{x}_d\right\|_2 + c}{\left\|\mathbf{x}_r\right\|_2 + \left\|\mathbf{x}_d\right\|_2 + c}
	\end{aligned}
	\label{eq:amp}
\end{equation}
where $c$ is the same positive constant used in \eqref{eq:ang} . It is easy to see that $0<\beta<1$. 

We propose a function $S(\mathbf{p}_r, \mathbf{p}_d)$ that measures the perceptual quality of $\mathbf{p}_d$ w.r.t~$\mathbf{p}_r$ as follows:
\begin{equation}
	\begin{aligned}
		S(\mathbf{p}_r, \mathbf{p}_d) = \alpha(\mathbf{p}_r, \mathbf{p}_d)\beta(\mathbf{p}_r, \mathbf{p}_d)
	\end{aligned}
	\label{eq:spatch}
\end{equation}
Let $S(\mathbf{p}_r^i, \mathbf{p}_d^i)$ be the quality measure between the $i$th pair of salient patches i.e.\ $(\mathbf{p}_r^i, \mathbf{p}_d^i)$. The proposed global image quality $\mathrm{SPARQ}(I_{ref}, I_{dis})$ is computed by averaging over all $q$ salient patches. 
\begin{equation}
	\begin{aligned}
		\mathrm{SPARQ}(I_{ref}, I_{dis}) = \frac{1}{q} \sum_{i=1}^q S(\mathbf{p}_r^i, \mathbf{p}_d^i)
	\end{aligned}
\end{equation}
\emph{Remarks:} 
\begin{itemize}
	\item The SPARQ index is bounded: $0< \mathrm{SPARQ}< 1$; it is always non-negative since each of its components is non-negative. 
	\item The highest value of SPARQ is attained when $I_{ref} = I_{dis}$. 
	\item The index is \emph{not} symmetric i.e.\ $\mathrm{SPARQ}(I_{ref}, I_{dis})\neq \mathrm{SPARQ}(I_{dis}, I_{ref})$. This is because the dictionary $\mathbf{\Phi}$ is trained on the reference image only. For the purpose of full-reference image quality assessment, where clear information about the reference image is available, this is not an issue. Nevertheless, symmetry can be easily achieved by repeating the quality estimation stage with a dictionary trained on the distorted image and averaging the resulting quality scores obtained using the two dictionaries. Our experiments show that this step has little or no significance on the performance of the SPARQ index. 
\end{itemize}
\section{Experimental Validation}
\label{sec:experiments}
This section presents a critical evaluation of the proposed metric on six publicly available image databases whose subjective quality ratings are available. These databases exhibit a variety of distortions such as compression artifacts, blurring, flicker noise, wireless artifacts, etc. The performance of an objective quality assessment metric is evaluated by comparing its results to the subjective scores. Following an evaluation methodology suggested by the video quality expert group (VQEG) \cite{vqeg}, this comparison is made by computing correlation coefficients and differences between the subjective and the objective scores. The objective scores of the SPARQ index and those of six existing image quality assessment metrics are compared to the subjective ratings on each dataset. The six image quality assessment metrics are: PSNR, SSIM \cite{ssim}, PHVS-M \cite{phvsm}, IFC \cite{ifc}, VIF \cite{vif}, and VSNR \cite{vsnr}. The existing quality metrics are compared to the SPARQ index on the basis of their closeness to the subjective scores. The SPARQ index consistently exhibits high correlation with the subjective ratings on all datasets and performs better or at par with the state-of-the-art.
\subsection{The databases}
\label{subsec:databases}
A brief description of each of the six datasets used in this work is provided below.

The \emph{LIVE} database \cite{ssim,live} contains $779$ distorted images created from $29$ original color images. Each distorted image exhibits one of the five types of distortions: JPEG2000 compression (JP2K), JPEG compression (JPEG), additive white gaussian noise (AWGN), Gaussian blur and fastfading channel distortion of JPEG2000 compressed bitstreams.

The \emph{Cornell-A57} dataset \cite{vsnr,a57} consists of $54$ distorted images created from $3$ original grayscale images. The images are subject to the following $6$ types of distortions: JPEG compression, JP2K compression, AWGN, Gaussian blur, JPEG2000 compression with dynamic contrast-based quantization algorithm, and uniform quantization of LH subbands of a 5-level discrete wavelet transform at all scales.

The \emph{CSIQ} database \cite{csiq} has $30$ original images which were used to create $866$ distorted images. The $6$ distortion types (at four to five distortion levels) include JPEG compression, JP2K compression, global contrast decrements, AWGN, and Gaussian blurring.

The \emph{TID} database \cite{tid} is so far the largest subject-rated image dataset for quality evaluation. It has $1700$ images generated from $25$ reference images with $17$ distortion types at four distortion levels. The distortion types are: AWGN, additive noise in color components, spatially correlated noise, masked noise, high frequency noise, impulse noise, quantization noise, Gaussian blur, image denoising, JPEG compression, JP2K compression, JPEG transmission errors, JP2K transmission errors, non-eccentricity pattern noise, local block-wise distortions of different intensity, mean shift, and contrast change.

The \emph{MICT-Toyoma} database \cite{toyoma} contains $168$ distorted images created from $14$ reference images. The images exhibit $2$ types of distortions: JPEG and JP2K compression.

The \emph{WIQ} database \cite{wiq,wiq1} consists of $80$ distorted images generated from $7$ reference images. The images exhibit wireless imaging artifacts which are not considered in other datasets. Due to the complex nature of a wireless communication channel, the images contain more than one artifacts.
\begin{figure}[tb]
	\centering
		\includegraphics[width=1.0\linewidth, trim= 2cm 0cm 2cm 1cm, clip=true]{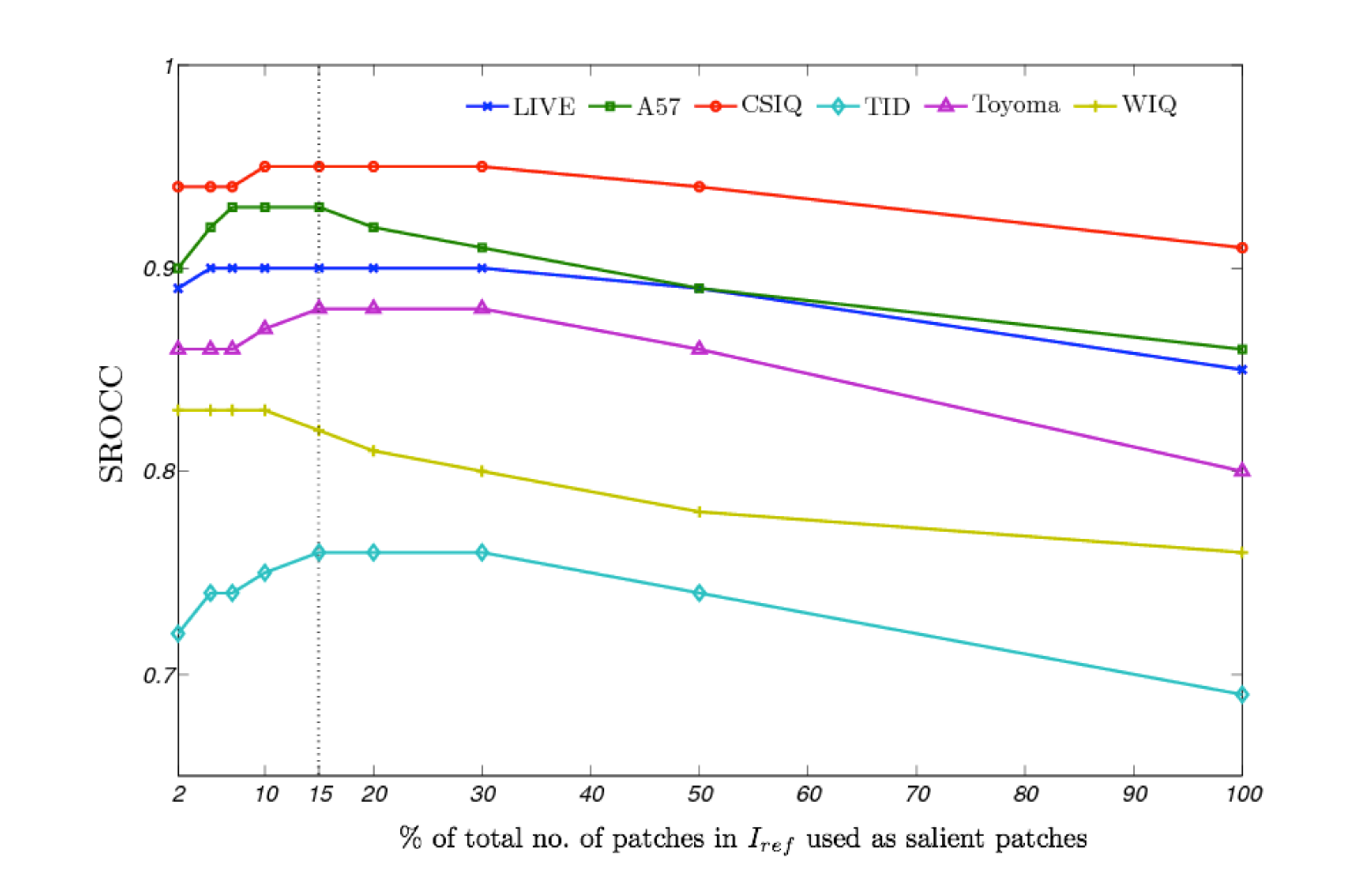}
	\caption{Performance of the SPARQ index (correlation with subjective scores measured in terms of SROCC) varies with the percentage of high-entropy patches used in the quality estimation process.}
	\label{fig:patchesSROCC}
\end{figure}

\subsection{Parameter settings}
\label{subsec:implementation}
Before computing the SPARQ index, two preprocessing steps are executed: (1) every color image in each dataset is converted to grayscale image, and (2) each image is downsampled by a factor $F$ so as to account for the viewing condition. The value of $F$ is obtained by using the following empirical formula \cite{ssim}.
\begin{equation}
	F = \mathrm{max}(1,\mathrm{round}(g/256))
	\label{eq:scale}
\end{equation}
where $g = \mathrm{min}\left(\# \mathrm{rows~in}~I_{ref}, \# \mathrm{columns~in}~I_{ref}\right)$.  

The computation of the SPARQ index is divided into a training phase and a quality estimation phase. In the training phase, there are $4$ parameters to be set: 
\begin{itemize}
	\item $\sqrt{n}$ : the patch size 
	\item $k$ : the number of patches to be extracted from a reference image for training the dictionary 
	\item $m$ : the number of basis vectors in the dictionary
	\item $\tau$ : the sparsity constraint
\end{itemize}
Unfortunately, there is no theoretical guidelines to determine the values of these parameter, so we rely on previous work and empirical methods. A patch size of $\sqrt{n}\times \sqrt{n} = 11 \times 11$ is used following the patch-size specification of SSIM \cite{ssim}. A collection of as large as $k = 3000$ patches are extracted \emph{randomly} from every reference image to train its corresponding dictionary. We set the overcompleteness factor ($m/n$) to $2$ which yields $m=242$. It has been shown that for low overcompleteness factor, sparse representations are stable in the presence of noise \cite{wohlberg}. The value of $\tau$ is set to $12$ which is approximately $10\%$ of the dimensionality of the input vectors.

In the quality estimation phase, we need $2$ additional parameters: 
\begin{itemize}
	\item $c$ : the stabilizing constant in \eqref{eq:ang} and \eqref{eq:amp} 
	\item $q$ : the number of salient patches
\end{itemize}
The constant $c$ is chosen to have a very small value, $c = 0.01$, so as to have minimal influence on the quality score. 

The value of $q$ is determined empirically. For each database, the number of salient patches, $q$, is varied and the performance of SPARQ is measured in terms of the correlation between its scores and the subjective scores. This is presented in Fig. \ref{fig:patchesSROCC} where the Spearman's Rank Correlation Coefficient (SROCC) is plotted against $q$. The value of $q$ is varied from $2\%$ to $100\%$ of $N$ where $N$ is the total number of patches (one around  each pixel) in $I_{ref}$ or $I_{dis}$. In five out of the six datasets, the best performance of the SPARQ index is observed when $q=0.15N$ i.e.~$15\%$ of $N$. Also notice that, when all patches in $I_{ref}$ are used, the performance of the SPARQ index degrades. This confirms our assumption that only the visually important areas are useful for quality assessment. For all datasets, we use the same parameter values.
\subsection{Evaluation methodology}
\label{subsec:discussion}
The results of an objective image quality assessment metric is compared with the subjective scores using a set of evaluation measures suggested by the video quality expert group (VQEG)\cite{vqeg}. These evaluation measures are - the Spearman's rank order correlation coefficient (SROCC), the Kendall's rank order correlation coefficient (KROCC), the Pearson linear correlation coefficient (CC), mean absolute error (MAE) and root mean squared error (RMS). 
The SROCC and KROCC are used to measure the \emph{prediction monotonicity}, while CC, MAE and RMS measure the \emph{prediction accuracy} of the objective scores. In order to compute CC, MAE and RMS, a five-parameter logistic function (refer to \eqref{eq:mapping} and \eqref{eq:logistic}) is fitted to the objective scores. A particular objective score, $s$, is mapped to a new score, $Q(s)$ using a non-linear mapping function $Q(\cdot)$ which is defined as follows.
 \begin{equation}
 	Q(s) = \gamma_1\mathrm{logistic}(\gamma_2,(s-\gamma_3))+s\gamma_4+\gamma_5
	\label{eq:mapping}
\end{equation}
\begin{equation}
 	\mathrm{logistic}(\sigma,s) = \frac{1}{2} - \frac{1}{1+\mathrm{exp}(\sigma, s)}
 	\label{eq:logistic}
 \end{equation}
%

\begin{table}[tb]
	\centering
	\caption{Performance of SPARQ Index on various datasets for different distortion types}
		\begin{tabular}{|c|c c c c c|}
			\hline
			\multicolumn{6}{|c|}{LIVE database} \tabularnewline
			\hline
			 				                &  SROCC	& KROCC	&  CC &   MAE &  RMS \tabularnewline
			\hline
			JPEG		                & 0.967 &	0.844 & 0.974  & 5.504	& 7.207 \tabularnewline
			
			JP2K			              & 0.939 & 0.781 & 0.946	& 6.201	& 8.164 \tabularnewline
			
			AWGN			             & 0.975	& 0.864 & 0.979 & 4.498	& 5.632 \tabularnewline
			
			Blurring           & 0.932 & 0.775 & 0.927 & 5.123	& 6.923 \tabularnewline
			
			Fastfading              & 0.904	& 0.747 & 0.905	& 9.129	& 12.134 \tabularnewline
			
			%
			\hline \hline
			\multicolumn{6}{|c|}{A57 database} \tabularnewline
			\hline
			 			 				                &  SROCC	& KROCC	&  CC &   MAE &  RMS \tabularnewline
			\hline
			JPEG		          						& 0.968	& 0.894 & 0.968 & 0.054 & 0.064 \tabularnewline
			
			JP2K			        						& 0.973	& 0.917 & 0.943 & 0.069 & 0.074 \tabularnewline
			
			AWGN				       						& 0.967	& 0.889 & 0.965 & 0.029 & 0.034 \tabularnewline
			
			Blurring     									& 0.912	& 0.772 & 0.953 & 0.046 & 0.060  \tabularnewline
			
			Quantized      						& 0.983 & 0.944 & 0.977 & 0.042 & 0.051 \tabularnewline
			
			JP2K-DCQ          						& 0.955	& 0.878 & 0.984 & 0.029 & 0.038 \tabularnewline
			
			%
			\hline  \hline
			\multicolumn{6}{|c|}{CSIQ database} \tabularnewline
			\hline
			 			 				                &  SROCC	& KROCC	&  CC &   MAE &  RMS \tabularnewline
			\hline
			JPEG		                & 0.972	& 0.858 & 0.986	& 0.041	& 0.054 \tabularnewline
			
			JP2K			              & 0.974	& 0.872 & 0.979	& 0.051	& 0.065 \tabularnewline
			
			AWGN             			& 0.952	& 0.811 & 0.939	& 0.045	& 0.058 \tabularnewline
			
			Blurring           & 0.975	& 0.865 & 0.978	& 0.048	& 0.060 \tabularnewline
			
			Contrast                & 0.911	& 0.761 & 0.916	& 0.050	& 0.067 \tabularnewline
			
			Pink noise              & 0.947	& 0.794 & 0.946	& 0.060	& 0.073 \tabularnewline
			
			%
			\hline \hline
			\multicolumn{6}{|c|}{TID database} \tabularnewline
			\hline
			 			 				                &  SROCC	& KROCC	&  CC &   MAE &  RMS \tabularnewline
			\hline
			JPEG		                			& 0.917	& 0.7268 & 0.951	& 0.403	& 0.526 \tabularnewline
			
			JP2K			              			& 0.963	& 0.8323 & 0.970	& 0.367	& 0.470 \tabularnewline
			
			AWGN             			& 0.756	& 0.5461 & 0.740	& 0.316	& 0.410 \tabularnewline
			
			Blurring                			& 0.946	& 0.7981 & 0.940	& 0.301	& 0.401 \tabularnewline
			
			Contrast                			& 0.375	& 0.2311 & 0.441	& 0.986	& 1.100 \tabularnewline
			
			JPEG trans        				& 0.820	& 0.6102 & 0.838	& 0.580	& 0.711 \tabularnewline
			
			JP2K trans       				& 0.807	& 0.6089 & 0.809	& 0.378	& 0.473 \tabularnewline
			
			Color noise      							& 0.788	& 0.5923 & 0.787	& 0.240	& 0.315 \tabularnewline
			
			Corr noise              			& 0.768	& 0.5758 & 0.760	& 0.309	& 0.406 \tabularnewline
			
			Mask noise              			& 0.856	& 0.6601 & 0.877	& 0.231	& 0.286 \tabularnewline
			
			Hi frq noise          				& 0.890	& 0.6889 & 0.901	& 0.297	& 0.404 \tabularnewline
			
			Impluse                 			& 0.789	& 0.5918 & 0.769	& 0.257	& 0.327 \tabularnewline
			
			Quantization            			& 0.814	& 0.6275 & 0.811	& 0.374	& 0.481 \tabularnewline
			
			Denoising               			& 0.928	& 0.7702 & 0.939	& 0.429	& 0.549 \tabularnewline
			
			Pattern noise            			& 0.724	& 0.5287 & 0.705 &	0.538	& 0.740 \tabularnewline
			
			Block wise              			& 0.724	& 0.5321 & 0.755	& 0.350	& 0.434 \tabularnewline
			
			Mean shift              			& 0.591	& 0.4147 & 0.653 &	0.358 &	0.436 \tabularnewline
			
			%
			\hline \hline
			\multicolumn{6}{|c|}{MICT database} \tabularnewline
			\hline
			 			 				                &  SROCC	& KROCC	&  CC &   MAE &  RMS \tabularnewline
			\hline
			JPEG		                & 0.877	& 0.691 & 0.883 &	0.462 &	0.580 \tabularnewline
			
			JP2K		                & 0.928 & 0.766 & 0.931	& 0.364	& 0.461 \tabularnewline
			
			%
			\hline \hline
			\multicolumn{6}{|c|}{WIQ database} \tabularnewline
			\hline
			 			 				                &  SROCC	& KROCC	&  CC &   MAE &  RMS \tabularnewline
			\hline
			Artifacts 1		& 0.822 & 0.640 & 0.823 & 10.899 & 12.929 \tabularnewline
			
			Artifacts 2		& 0.836 & 0.688 & 0.894 & 7.437 & 10.291 \tabularnewline
		
			\hline
		\end{tabular}
		\label{tab:alldistortions}
\end{table}

A MATLAB function called \emph{fminunc} is used for fitting. CC, MAE and RMS values are computed after the above non-linear mapping between the subjective and objective scores. Note that, SROCC and KROCC are non-parametric rank correlation metrics and are independent of any nonlinear mapping between the subjective and the objective scores. For details of the evaluation methodology please see \cite{vqeg, iwssim, vif}. A good image quality assessment metric is expected to have high SROCC, KROCC and CC scores, and low MAE and RMS values.

The performance of SPARQ is compared with those of PSNR, SSIM, PHVS-M, IFC, VIF and VSNR on the basis of their correlation and differences with the subjective ratings. PSNR is used as a baseline method. PHVS-M and VSNR are the HVS-based IQA metrics while SSIM, IFC, VIF and SPARQ are visual fidelity-based metrics. For the implementation of SSIM, PHVS-M, IFC, VIF and VSNR, we have used the original MATLAB codes provided by the respective authors. The parameters of each of these methods are set to their default values as suggested in the original references.
\begin{table*}[t]
	\caption{Overall performance comparison of IQA algorithms}
	\centering
	\begin{tabular}{|c|c c c c c c c|}
		\hline
		\multicolumn{8}{|c|}{\textbf{\emph{SROCC-based comparison}}} \tabularnewline
		\hline
		Dataset & PSNR & SSIM \cite{ssim} & PHVSM \cite{phvsm}	& IFC \cite{ifc}  & VIF \cite{vif} & VSNR \cite{vsnr} & SPARQ \tabularnewline
		\hline																																																															
		LIVE      & 0.875	& \bf0.947		   & 0.922			          & 0.926	          & \bf 0.963	  & 0.912	    & 0.930		\tabularnewline
			
		A57  	   & 0.598	& 0.806			   & 0.896			      & 0.318	          & 0.622		  & \bf 0.935     & \bf0.931	  \tabularnewline
			
		CSIQ   & 0.800	& 0.858			 & 0.822			 & 0.767	          & \bf0.919	         & 0.809 	& \bf 0.951 \tabularnewline
			
		TID       & 0.552	& \bf 0.773		     & 0.561			        & 0.622	          & 0.749			     & 0.704 		& \bf0.759 \tabularnewline
			
		MICT    & 0.613	& 0.875	       & 0.848			        & 0.835     			& \bf 0.907	     & 0.860 			&	\bf0.879 \tabularnewline
			
		WIQ	& 0.626	& \bf0.758	           & 0.757		          & 0.716	          & 0.692		       & 0.656 		& \bf 0.822 \tabularnewline
		\hline 
		\multicolumn{8}{|c|}{\emph{performance over all datasets}} \tabularnewline
		\hline
		Direct average	 & 0.677	& \bf0.837 & 0.801 & 0.697 & 0.809 & 0.813 & \textbf{0.878} \tabularnewline
			
		Weighted average & 0.685 & 0.838 & 0.722 & 0.729 & \bf0.839 & 0.783 & \textbf{0.851} \tabularnewline
		\hline \hline
      	\multicolumn{8}{|c|}{\textbf{\emph{CC-based comparison}}} \tabularnewline
		\hline
		Dataset	& PSNR	& SSIM \cite{ssim} & PHVSM \cite{phvsm}	& IFC \cite{ifc}  & VIF \cite{vif} & VSNR \cite{vsnr} & SPARQ   	\tabularnewline
		\hline
		LIVE         & 0.860	& \bf0.941	     & 0.917			        & 0.853	          & \bf 0.944			 & 0.917		& 0.929	 \tabularnewline
			
		A57        	& 0.628	& 0.802			       & 0.875			      & 0.372	     & 0.614			 & \bf 0.914 		& \bf 0.936 \tabularnewline
			
		CSIQ     	  & 0.746	& 0.758	       & 0.772			        & 0.821	          & \bf0.927		& 0.735 	              & \bf 0.947 \tabularnewline
			
	TID    	& 0.519	& 0.727			       & 0.552			        & 0.660	          & \bf 0.809			 & 0.682 			& \bf 0.788	
			\tabularnewline
			
	MICT      	& 0.632	& 0.705	   & 0.839			        & 0.833     			& \bf 0.902	     & 0.855 	&	\bf0.883     \tabularnewline
			
	WIQ		& 0.639	& 0.640	           & 0.749		          & 0.705	          & 0.730		   & \bf0.763 		& \bf 0.794 \tabularnewline
	\hline 
	\multicolumn{8}{|c|}{\emph{performance over all datasets}} \tabularnewline
	\hline 
			Direct average  & 0.687 & 0.762 & 0.784 & 0.707 & \bf0.821 & 0.811&\textbf{0.879} \tabularnewline
			
			Weighted average& 0.657 & 0.778 & 0.704 & 0.744 & \textbf{0.865} & 0.758 & \textbf{0.862} \tabularnewline
	\hline\hline
      \multicolumn{8}{|c|}{\textbf{\emph{RMS-based comparison}}} \tabularnewline
	\hline
	Dataset	& PSNR	& SSIM \cite{ssim} & PHVSM \cite{phvsm}	& IFC \cite{ifc}  & VIF \cite{vif} & VSNR \cite{vsnr} & SPARQ 	\tabularnewline
	\hline
	LIVE     	& 13.990 	& \bf9.985	               & 10.892			        & 14.263	   & \bf 9.240    & 10.772	& 10.118 \tabularnewline
			
	A57          & 0.191	 	& 0.147			 & 0.119			         & 0.223	          & 0.194	      & \bf0.099 		& \bf 0.086 \tabularnewline
			
	CSIQ      	& 0.175	 	& 0.171			 & 0.167	    & 0.150	          & \bf 0.098	      & 0.178 		       & \bf 0.084 	\tabularnewline
			
	TID        	& 1.147	 	& 0.921			 & 1.119		   & 1.008	          & \bf 0.789	      & 0.981 			& \bf0.805		\tabularnewline
			
 MICT         	& 0.969	 	& 0.887			 & 0.680	      & 0.692     	   & \bf 0.540	      & 0.648 			& \bf0.588 \tabularnewline
			
WIQ			& 15.426       & 17.595	        & 15.185		     & 16.252	   & 15.653		& \bf14.809 		& \bf 13.906 	\tabularnewline
			\hline 
			\multicolumn{8}{|c|}{\emph{performance over all datasets}} \tabularnewline
			\hline
			Direct average & 5.316	       & 4.951	        & 4.694					      &  5.431                &  \bf4.419		& 4.581			& \bf 4.264 \tabularnewline
		
			Weighted average & 3.950	& 3.035	               & 3.254					& 3.944           &\bf  2.736            &   3.156              & \bf 2.889\tabularnewline
			\hline
			\end{tabular}
	\label{tab:comparison1}		
\end{table*}	
\subsection{Performance comparison} 
Table \ref{tab:alldistortions} lists the performance of SPARQ when compared to the subjective ratings on each database, for each distortion type separately. The high correlation values obtained in most of the cases show that SPARQ works well for a variety of distortion types. 

Table \ref{tab:comparison1} compares the overall performance of SPARQ with the state-of-the-art image quality assessment metrics in terms of SROCC, CC and RMS. KROCC and MAE are left out since they reflect the same performance trend as SROCC and RMS, respectively. In order to provide the big picture, the average SROCC, CC and RMS values are computed over all six datasets. The average values are computed for two cases: in the first case the (SROCC or CC or RMS) values are directly averaged and in the second case the values are weighted by the size of the databases. The weight for a particular database is the number of distorted images it contains, e.g.~779 for LIVE and 54 for A57. In each case, the best two results are printed in boldface.\\
From Table \ref{tab:comparison1}, we see that VIF is the closest competitor of SPARQ. Hence we performed a detailed comparison between SPARQ and VIF by comparing their performances for each distortion types separately. This comparison is presented in Table \ref{tab:vifVssparq}.

\noindent\emph{Remarks}:
\begin{itemize}
	\item SPARQ clearly outperforms PSNR, PHVS-M and IFC on all datasets. 
	\item SPARQ outperforms VSNR on $5$ out of $6$ datasets. On the A57 dataset, SPARQ's performances is comparable to VSNR in terms of SROCC, but it is better than VSNR in terms of CC and RMS values. (see Table \ref{tab:comparison1})
	\item In terms of overall performance, SPARQ is \emph{better or comparable to VIF}. However, the performance of VIF varies much (e.g.~SROCC = 0.963 on LIVE but SROCC = 0.622 on A57) over the datasets, while SPARQ's performance is \emph{more consistent}.	
	\item The distortion-specific performance comparison in Table \ref{tab:vifVssparq} shows that SPARQ performs \emph{better than VIF}. 
	\item The WIQ dataset is the only dataset that contains more than one artifacts due to the nature of wireless imaging. Notice that, SPARQ handles such complex artifacts much better than any other metric. This indicates the potential of SPARQ index to be used in complex practical systems where degradation of images is likely to be caused by more than one factors.
\end{itemize}
\begin{table*}[tb]
	\centering
	\caption{Distortion-specific Performance Comparison Between VIF and SPARQ in Terms of CC}
		\begin{tabular}{|c|c| c c ||c|c|c c|}
			\hline
			 Distortion & Database	 & SPARQ       &   VIF \cite{vif}  & Distortion & Database				    & SPARQ       &   VIF \cite{vif}\tabularnewline
			\hline \hline
			 JPEG 	&LIVE	     	   & 0.974  		& \bf 0.987 	 & JP2K & LIVE		           & 0.946			& \bf 0.977\tabularnewline
			
						& A57		         		    & \bf 0.968		& 0.950 &  & A57		        	& \bf 0.943 		& 0.865 		 \tabularnewline
			
			 & CSIQ		          	   & \bf 0.986		& 0.985 	&  & CSIQ		          & 0.979			& \bf 0.982	\tabularnewline
			
			 &TID            			  & \bf 0.951			& 0.911 & & TID			      & 0.970				& \bf 0.976	     \tabularnewline
			
			&	MICT	               	  & 0.883 			& \bf 0.892 & & MICT	          & 0.931	&\bf 0.949\tabularnewline
			\hline
			AWGN & LIVE			        & 0.979 		& \bf 0.990	& Blur &LIVE           & 0.927 		& \bf 0.974 \tabularnewline
			
			 & A57			& \bf 0.965 		& 0.881 		&  & A57     	 & \bf 0.953 		& 0.945\tabularnewline
			
			 & CSIQ            	 & 0.939			& \bf 0.952  & &CSIQ           & \bf 0.978		& 0.966 \tabularnewline
			
			 & TID           	 & \bf 0.740			& 0.686 & 	&TID             & 0.940				& \bf 0.952 \tabularnewline

			\hline
			 Quantization & A57     	& \bf 0.977 		& 0.842  	&	Contrast change & CSIQ             & \bf 0.916		& 0.915	\tabularnewline
			  & TID     & \bf 0.811	& 0.374 &  & TID               & 0.441				& \bf 0.945\tabularnewline
			\hline
			
			Fastfading & LIVE              & 0.905			& \bf 0.956   & JP2K-DCQ & A57              & \bf 0.984 		& 0.967		 \tabularnewline
			\hline
			
			 Pink noise & CSIQ              & 0.946			& \bf 0.959	& JPEG transmission & TID          & 0.838				& \bf 0.873\tabularnewline
			\hline
			
			JP2K transmission &TID      	  & \bf 0.809			& 0.770	&  Color noise &TID     	  & \bf 0.787			& 0.618\tabularnewline
			\hline
			
			Correlated noise & TID          & \bf 0.760			& 0.147	& Mask noise & TID         & \bf 0.877	& 0.685 \tabularnewline
			\hline
			Hi Frequency noise & TID          & \bf 0.901	& 0.885	& 	Impulse noise & TID            & 0.769	& \bf 0.831 \tabularnewline
			\hline
			Denoising & TID           & 0.939	& \bf 0.973	& Pattern noise & TID      & \bf 0.705 &	0.686 \tabularnewline

			\hline
			
			 Blockwise distortion & TID         & 0.755	& \bf 0.828 & Mean shift & TID        & \bf 0.653 &	0.540  \tabularnewline
			\hline
			Wireless artifact 1 & WIQ	& \bf0.823  &  0.762 & Wireless artifact 2 & WIQ	& \bf 0.894 & 0.729\tabularnewline
			\hline\hline
			 \multicolumn{8}{|c|}{SPARQ is better in \textbf{21} cases while VIF is better in 17 cases} \tabularnewline
			 \hline
		\end{tabular}
		\label{tab:vifVssparq}
\end{table*}
%
%
\subsubsection{Computational complexity}
In order to compute the SPARQ index, the two steps that require the bulk of computation are (i) the dictionary learning step in the training phase and (ii) the sparse coding step in the quality estimation phase. The computational load of the dictionary learning step in turn is dominated by the sparse coding step performed as part of the learning process. Hence, it is the sparse coding step that we should be concerned with. 

Our implementation uses an efficient sparse coding algorithm called the \emph{Batch-OMP} \cite{efficientKSVD}. Its computational complexity is $\mathcal{O}(nm\tau)$ per training signal, where the dictionary dimension is $n\times m$ and $\tau$ is the sparsity constraint and $\tau<<m$ \cite{efficientKSVD}. 

To give an idea of the computation time, a basic Matlab implementation (using a computer with Intel Q9400 processor at 2.66 GHz) takes about $3.4$ seconds to learn a dictionary of size $121\times 242$ with $\tau=12$ using $k = 3000$ training samples extracted from an image of dimension $256 \times 256$. The quality estimation takes about $0.9$ sec. The total time required to perform quality evaluation on the LIVE dataset is $779.7$ secs (learning: $29\times3.4$ secs + quality estimation: $779\times0.9$ secs) i.e.\ $\sim 1$ sec processing time per distorted image. 
Like any method involving training, the dictionary learning step can be performed offline and the dictionaries can be precomputed. 

%
\subsubsection{Limitations of SPARQ}
Due to its dependence on sparse coding, SPARQ is computationally demanding. We are hopeful that with further progress in this area faster algorithms will be available in near future. 

The SPARQ index works on grayscale images and thus is blind to the degradations in the color components. Like most of the existing IQA metrics, SPARQ relies on fidelity to quantify perceptual quality where fidelity is one of the several factors in determining the perceptual quality \cite{fidelityvsquality}.

\section{Conclusion}
\label{sec:conclusion}
In this paper, we develop a new full-reference image quality assessment metric, namely the SPARQ index. This metric relies on learning an overcomplete dictionary from the reference image. The basis elements of this dictionary are learnt using a sparse optimization approach and they resemble the receptive field of simple cells in the primary visual cortex. The SPARQ index measures the structural fidelity between the reference and the distorted image in order to quantify the visual quality of the distorted image. 

The SPARQ index is shown to be consistently performing better or comparable to the state-of-the-art. The success of SPARQ can be attributed to the new framework that can extract \emph{perceptually meaningful} structural information by modeling the response of the primary visual cortex to the stimuli. 

The SPARQ index can be easily applied to other problems involving similarity measurement such as clustering. Because of its generic data-dependent approach, SPARQ is also suitable (may require minor modifications) for various datatypes including images, videos and audio signals.

The SPARQ index can be improved in several ways. Possible directions include combining SPARQ with various pooling strategies, learning multiscale dictionaries, using more efficient sparse solvers and extending it to work for color images and videos.

\IEEEtriggeratref{33}


\bibliographystyle{IEEEtran}
\bibliography{refs}
%
%
%



\newpage





\end{document}